\documentclass[10pt,twocolumn,letterpaper]{article}

\usepackage{cvpr}
\usepackage{times}
\usepackage{epsfig}
\usepackage{graphicx}
\usepackage{amsmath}
\usepackage{amssymb}
\usepackage{caption}
\usepackage{amsmath}
\usepackage[font=small,skip=3pt]{caption}
\usepackage{authblk}

\usepackage{caption}
\usepackage[pagebackref=true,breaklinks=true,colorlinks,bookmarks=false]{hyperref}

 \cvprfinalcopy 

\def\cvprPaperID{1321} 

\ifcvprfinal\fi
\begin{document}

\title{DoubleFusion: Real-time Capture of Human Performances with Inner Body Shapes from a Single Depth Sensor\vspace{-0.6cm}}

\author{
	Tao Yu\textsuperscript{1,2},
	Zerong Zheng\textsuperscript{1},
	Kaiwen Guo\textsuperscript{1,3},
	Jianhui Zhao\textsuperscript{2},
	Qionghai Dai\textsuperscript{1},
	\authorcr
	Hao Li\textsuperscript{4},\quad
	Gerard Pons-Moll\textsuperscript{5},\quad
	Yebin Liu\textsuperscript{1,6}
	\\
	\vspace{-0.1cm}
	\textsuperscript{1}Tsinghua University, Beijing, China
	\quad
	\textsuperscript{2}Beihang University, Beijing, China
	\quad
	\textsuperscript{3}Google Inc
	\authorcr
	\textsuperscript{4}University of Southern California / USC Institute for Creative Technologies
	\authorcr
	\textsuperscript{5}{Max-Planck-Institute for Informatics, Saarland Informatics Campus}
	\authorcr
	\textsuperscript{6}{Beijing National Research Center for Information Science and Technology (BNRist)}
}

\maketitle

\begin{abstract}
We propose DoubleFusion, a new real-time system that combines volumetric dynamic reconstruction with data-driven template fitting to simultaneously reconstruct detailed geometry, non-rigid motion and the inner human body shape from a single depth camera. One of the key contributions of this method is a double layer representation consisting of a complete parametric body shape inside, and a gradually fused outer surface layer. A pre-defined node graph on the body surface parameterizes the non-rigid deformations near the body, and a free-form dynamically changing graph parameterizes the outer surface layer far from the body, which allows more general reconstruction. We further propose a joint motion tracking method based on the double layer representation to enable robust and fast motion tracking performance. Moreover, the inner body shape is optimized online and forced to fit inside the outer surface layer. Overall, our method enables increasingly denoised, detailed and complete surface reconstructions, fast motion tracking performance and plausible inner body shape reconstruction in real-time. In particular, experiments show improved fast motion tracking and loop closure performance on more challenging scenarios.

\end{abstract}

\section{Introduction}
\label{sec:intro}
\begin{figure}
    \centering
    \includegraphics[width=1.0\linewidth]{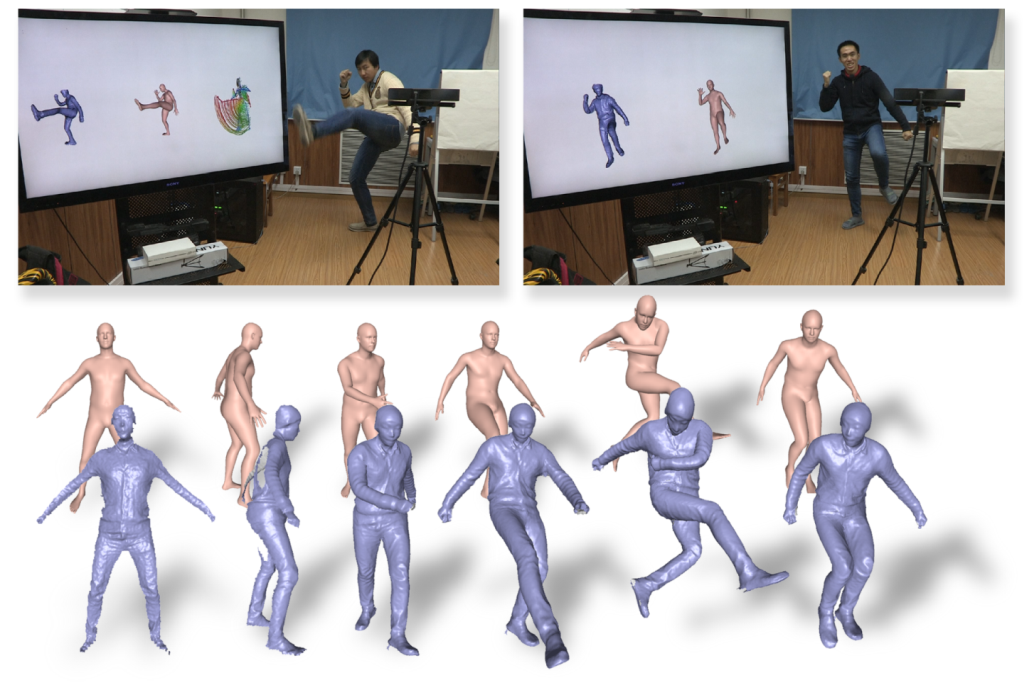}
    \caption{Our system and the real-time reconstructed results.}
    \label{fig:teaser}
\end{figure}
Human performance capture has been a challenging research topic in computer vision and computer graphics for decades. The goal is to reconstruct a temporally coherent representation of the dynamically deforming surface of human characters from videos. Although array based methods~\cite{Liu:2010:PMS:1749404.1749522,gall2009motion,bradley2008markerless,brox2010combined,ye2012performance,LiuGSDST13multiple,MustafaKGH15general,dou2016fusion4d,Vincent2017integration,Pons-Moll:Siggraph2017} using multiple video or depth cameras are well studied and have achieved high quality results, the expensive camera-array setups and controlled studios limit its application to a few technical experts. As depth cameras are increasingly popular in the consumer space (iPhoneX, Google Tango, etc.), the recent trend focuses on using more and more practical setups like a single depth camera~\cite{Zollhofer2014,guo2015robust,BogoBL015Detailed}. In particular, by combining non-rigid surface tracking and volumetric depth integration, DynamicFusion like approaches~\cite{newcombe2015dynamic,innmann2016volume,guo2017real,slavcheva2017cvpr} allow real-time dynamic scene reconstruction using a single depth camera without the requirement of pre-scanned model templates. Such systems are low cost, easy to set up and promising for popularization; however, they are still restricted to controlled slow motions. The challenges are occlusions (single view), computational resources (real-time), loop closure and no pre-scanned template model.

BodyFusion~\cite{tao2017BodyFusion} is the most recent work in the direction of single-view real-time dynamic reconstruction; It shows that regularizing non-rigid deformations with a skeleton is beneficial to capture human performances. However, since the human joints are too sparse and it only uses the gradually fused surface for tracking, it fails during fast motions, especially when the surface is not yet complete. Moreover, the skeleton embedding performance relies heavily on the initialization step and is fixed afterwards. Inaccurate skeleton embedding results in deteriorated tracking and deformation performance. 

For human performance capture, besides the skeleton, body shape is also a very strong prior since it is loop closed and complete. To fully take advantage of \emph{both human shape and pose motion prior}, we propose ``DoubleFusion'': a single-view and real-time dynamic surface reconstruction system that simultaneously reconstructs general cloth geometry and inner body shape. In addition, we make each layer benefit from each other. Based on the recent state-of-the-art body model SMPL~\cite{Loper2015SMPL}, we propose a double-layer surface representation consisting of an outer surface layer, and an inner body layer for reconstruction and depth registration.
The observed outer surface is gradually fused and deformed while the shape and pose parameters of the inner body layer are also gradually optimized to fit inside the outer surface. On one hand, the inner body layer is a complete model that allows to find enough correspondences, especially when only partial surface is obtained; in addition, it places a constraint on where to fuse the geometry of the outer surface. On the other hand, the gradually fused outer surface provides increasingly more constraints to update the body shape and pose online. The two layers are solved sequentially in real-time.

Overall, our proposed DoubleFusion system offers the new ability to simultaneously reconstruct the inner body shape and pose as well as the outer surface geometry and motion in real-time. This is achieved by using only a single depth camera, and without pre-scanning efforts. Compared to systems that only reconstruct the outer surface like BodyFusion~\cite{tao2017BodyFusion}, we demonstrate substantially improved performance in handling fast motions. In contrast to systems specialized to capture the inner body~\cite{BogoBL015Detailed}, our approach can handle people wearing casual clothing, and it works in real-time.
To enable the above advantages, we make the following technical contributions in this paper.

\begin{itemize}
\item{We propose the double-layer representation (Section \ref{sec:double}) for high quality and realtime human performance capture. We define the double node graph that contains an on-body node graph and a far-body node graph. The double node graph enables better leverage of the human shape and pose prior, while still maintaining the ability to handle surface deformations that are far from the inner body surface. The double-layer representation may also be used in other human performance capture setups like multiview systems.}
\item{Joint motion tracking (Section \ref{sec:tracking}). We introduce a method to jointly optimize for the pose of the inner body shape and the non-rigid deformation of the outer surface based on the double-layer representation. Feature correspondences on both the inner body shape and fused outer layer enable fast motion tracking performance and robust geometry reconstruction. }
\item{Volumetric shape-pose optimization of the inner layer (Section \ref{sec:volumetric_fusion_opt}). We fit the SMPL model parameters with the canonical model directly in the TSDF volume defined by the outer surface without searching correspondences. The optimized body shape and pose (skeleton embedding) in the canonical frame is beneficial for outer surface tracking.}
\end{itemize}

\section{Related Work} \label{sec:related}
In this work, we focus on capturing the dynamic geometry of human performer with detailed surface and personal body shape identity using a single depth sensor. The related methods can roughly divided into static template based, model-based and free-form reconstruction methods.

\textbf{Static template based dynamic reconstruction.}
For performance capture, some of the previous works leverage pre-scanned templates. Thus surface reconstruction is turned into a motion tracking and surface deformation problem. Vlasic \emph{et al.} \cite{vlasic2008articulated} and Gall \emph{et al.} \cite{gall2009motion} adopted a template with embedded skeleton driven by multi-view silhouettes and temporal feature constraints. Liu \emph{et al.} \cite{liu2011markerless} extended the method to handle multiple interacting performers. 
Some approaches \cite{taylor2012vitruvian,Pons-Moll:IJCV:2015} use a random forest to predict correspondences to a template, and use them to fit the template to the depth data.
Ye \emph{et al.} \cite{ye2012performance} considered the case of multiple Kinects input. Ye \emph{et al.} \cite{ye2014real} adopted a similar skinned model to estimate shape and pose parameters using a single-view depth camera in real-time. For this kind of template, in order to achieve accurate tracking, skeleton embedding is usually done manually.

Besides templates with an embedded skeleton, some works adopted template based non-rigid surface deformation. Li \emph{et al.} \cite{li2009robust} utilized embedded deformation graph in Sumner \emph{et al.} ~\cite{Sumner2007embededed} to parameterize the pre-scanned template to produce locally as-rigid-as-possible deformation. Guo \emph{et al.} \cite{guo2015robust} adopted an $\ell_0$ norm constraint to generate articulate motions without explicitly embedded skeleton. Zollh{\"o}fer \emph{et al.} \cite{Zollhofer2014} took advantage of massive parallelism of GPU to enable real-time performance of general non-rigid tracking.

For the aforementioned require scanning a template step before capturing people with different identities or even the same performer with various apparels.

\textbf{Model-based dynamic reconstruction.}
In addition to pre-scanned templates, many general body models have been proposed in the last decades. SCAPE~\cite{Anguelov2005scape} is one of the widely used model, it factorizes deformations into pose and shape components. SMPL~\cite{Loper2015SMPL} is a recent body model that represents shape and pose dependent deformations in an efficient linear formulation. Dyna \cite{PonsMoll2015Dyna} learned a low-dimensional subspace to represent soft-tissue deformations.

Many research works utilized these shape priors to enforce more general constraints to capture dynamic bodies. Chen \emph{et al.} \cite{chen2016realtime} adopted SCAPE to capture body motion using a single depth camera. Bogo \emph{et al.} \cite{BogoBL015Detailed} extended SCAPE to capture detailed body shape with appearance. Bogo \emph{et al.} \cite{Bogo2016keepsmpl} used SMPL to fit predicted 2D joint locations to estimate human shape and pose. However, neither SCAPE nor SMPL can represent arbitrary geometry of the performer wearing various apparels. In Zhang \emph{et al.} \cite{zhang2017detailed} they addressed this problem by estimating the inner shape and recovering surface details. Pons-Moll \emph{et al.} \cite{Pons-Moll:Siggraph2017} introduce ClothCap, which jointly estimates clothing geometry and body shape using separate meshes. In both \cite{zhang2017detailed} and \cite{Pons-Moll:Siggraph2017}, results are only shown for complete 4D scan sequences. Alldieck  \emph{et al.} \cite{thiemo2018} reconstruct detailed shape including clothing from a monocular RGB video but the approach is off-line.


\textbf{Free-form dynamic reconstruction.} Free-form capture does not assume any geometric prior. For general non-rigid scenes, motion and geometry are closely coupled. In order to fuse regions visible in the future into a complete geometry, the algorithm needs to estimate non-rigid motion accurately. On the other hand, one needs accurate geometry to estimate motion accurately. In the last decades, many methods have been proposed to address free-form capture: linear variational deformation \cite{liao2009modeling}, deformation graph \cite{li2008global}, subspace deformation \cite{wand2009efficient}, articulate deformation \cite{chang2009range,chang2011global} and  \cite{pekelny2008articulated}, 4D spatio-temporal surface \cite{mitra2007dynamic} and \cite{sussmuth2008dyn}, incompressible flows \cite{sharf2008space}, animation cartography \cite{tevs2012animation}, quasi-rigid motions \cite{li20133d} and directional field \cite{dou2013scanning}.

Only in recent years, free-form capture methods with real-time performance have been proposed. DynamicFusion \cite{newcombe2015dynamic} proposed a hierarchical node graph structure and an approximate direct GPU solver to enable capturing non-rigid scenes in real-time. Guo \emph{et al.} \cite{guo2017real} proposed a real-time pipeline that utilized shading information of dynamic scenes to improve non-rigid registration, meanwhile accurate temporal correspondences are used to estimate surface appearance. Innmann \emph{et al.} \cite{innmann2016volume} used SIFT features to improve tracking and Slavcheva \emph{et al.} \cite{slavcheva2017cvpr} proposed a killing constraint for regularization. However, neither of methods demonstrated full body performance capture with natural motions. Fusion4D \cite{dou2016fusion4d} setup a rig with 8 depth camera to capture dynamic scenes with challenging motions in real-time. BodyFusion \cite{tao2017BodyFusion}, utilizes skeleton priors for human body reconstruction, but cannot handle challenging fast motions and cannot infer inner body shape. 

\section{Overview}
\label{sec:overview}
\subsection{Double-layer Surface Representation}
\label{sec:double}
\begin{figure}
	\centering
	\includegraphics[width=1.0\linewidth]{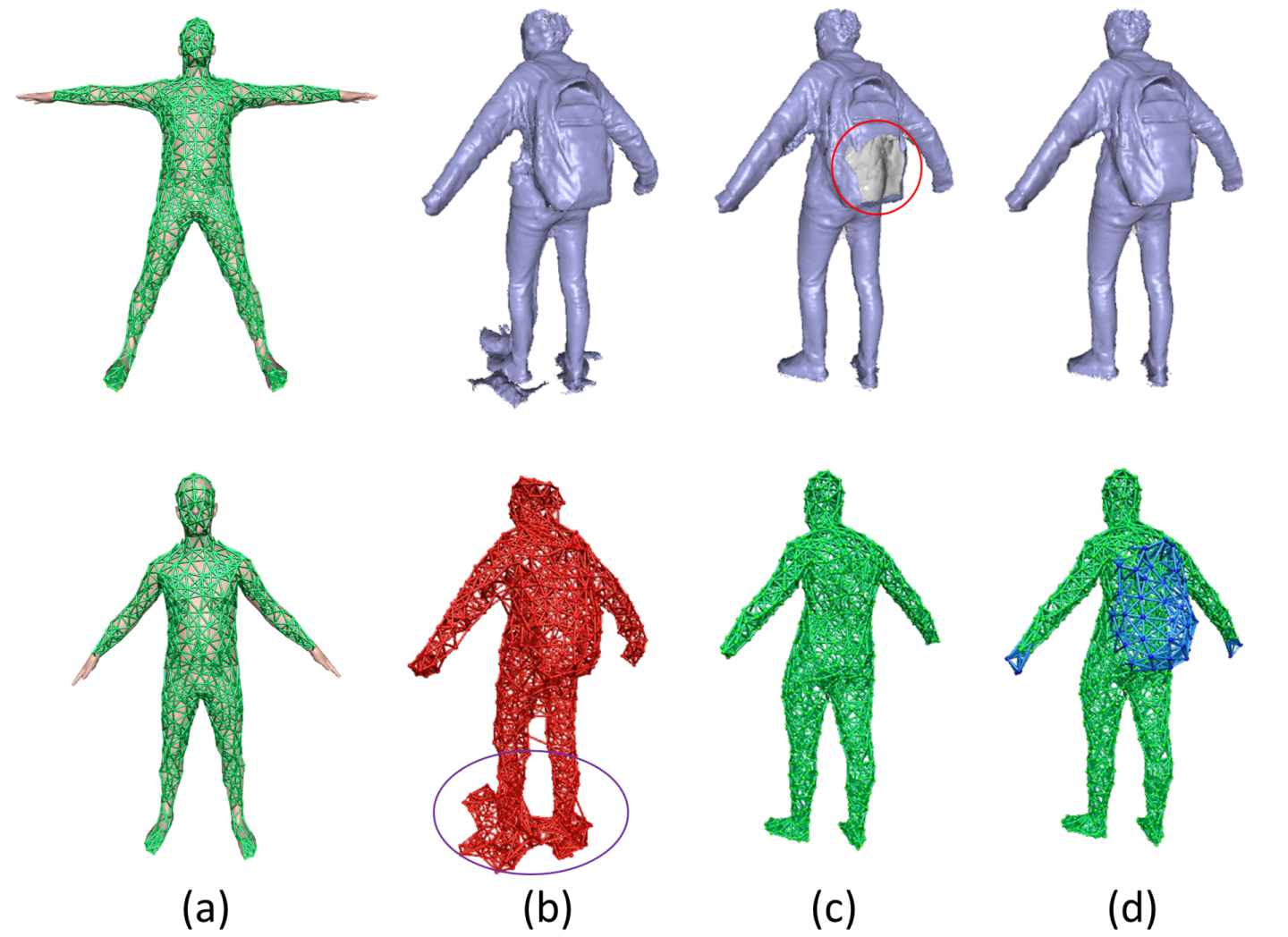}
	\caption{(a) Initialization of the on-body node graph. (b)(c)(d) Evaluation of the double node graph. The figure shows the geometry results and live node graph of (b) traditional free-form sampled node graph (red), (c) on-body node graph (green) only and (d) double node graph (with far-body nodes in blue). Note that we render the inner surface of the geometry in gray in (c)(top).}\vspace{6pt}
	\label{fig:eval_hieGraph}
\end{figure}

The input to DoubleFusion is a depth stream captured from a single consumer-level depth sensor and the output is a double-layer surface of the performer. The outer layer are observable surface regions, such as clothing, visible body parts (e.g. face, hair), while the inner layer is a parametric human shape and skeleton model based on the skinned multi-person linear model (SMPL)~\cite{Loper2015SMPL}. Similar to previous work~\cite{newcombe2015dynamic}, the motion of the outer surface is parametrized by a set of nodes. Every node deforms according to a rigid transformation. The~\emph{node graph} interconnects the nodes and constrain them to deform similarly. Unlike \cite{newcombe2015dynamic} that uniformly samples nodes on the newly fused surface, we pre-define an on-body node graph on the SMPL model, which provides a semantic and real prior to constrain non-rigid human motions. For example, it will prevent erroneous connections between body parts (e.g., connecting the legs). We uniformly sample on-body nodes and use geodesic distances to construct the predefined on-body node graph on the mean shape of SMPL model as shown in Fig.~\ref{fig:eval_hieGraph}(a)(top). The on-body nodes are inherently bound to skeleton joints in the SMPL model. Outer surface regions that are close to the inner body are bound to the on-body node graph. Deformations of regions far from the body cannot be accurately represented with the on-body graph. Hence, we additionally sample far-body nodes with a radius of $\delta=5cm$ on the newly fused far-body geometry. A vertex is labled as far-body when it is located further than $1.4\times\delta cm$ from its nearest on-body node, which helps to make sure the sampling scheme is robust against depth noise and tracking failures. The double node graph is shown in Fig.~\ref{fig:eval_hieGraph}(d)(bottom).  


\subsection{Inner Body Model: SMPL}
\label{subsec:smpl}
SMPL~\cite{Loper2015SMPL} is an efficient linear body model with $N=6890$ vertices. SMPL incorporates a skeleton with $K=24$ joints. Each joint has 3 rotational Degrees of Freedom (DoF). Including the global translation of the root joint, there are $3\times24+3=75$ pose parameters. Before posing, the body model $\bar{\mathbf{T}}$ deforms according to shape parameters $\boldsymbol{\beta}$ and pose parameters $\boldsymbol{\theta}$ to accommodate for different identities and non-rigid pose dependent deformations. Mathematically, the body shape $T(\boldsymbol{\beta},\boldsymbol{\theta})$ is morphed according to
\begin{equation} \label{eqn:smpl_template}
T(\boldsymbol{\beta},\boldsymbol{\theta}) = \bar{\mathbf{T}} + B_s(\boldsymbol{\beta}) + B_p(\boldsymbol{\theta})
\end{equation}
where $B_s(\boldsymbol{\beta})$ and $B_p(\boldsymbol{\theta})$ are vectors of vertex offsets, representing shape blendshapes and pose blendshapes respectively. The posed body model $M(\boldsymbol{\beta},\boldsymbol{\theta})$ is formulated as
\begin{equation} \label{eqn:smpl_pose}
M(\boldsymbol{\beta},\boldsymbol{\theta}) = W(T(\boldsymbol{\beta},\boldsymbol{\theta}), J(\boldsymbol{\beta}), {\boldsymbol{\theta}}, \mathcal{W})
\end{equation}
where $W(\cdot)$ is a general blend skinning function that takes the modified body shape $T(\boldsymbol{\beta},\boldsymbol{\theta})$, pose parameters $\boldsymbol{\theta}$, joint locations $J(\boldsymbol{\beta})$ and skinning weights $\mathcal{W}$, and returns posed vertices. Since all parameters were learned from data, the model produces very realistic shapes in different poses. We use the open sourced SMPL model with 10 shape blendshapes. See~\cite{Loper2015SMPL} for more details.

\subsection{Initialization}
\label{subsec:init}
During capture, we assume a fixed camera position and treat camera movement as global scene rigid motion. In the initialization step, we require the performer to start with a rough A-pose. For the first frame, we initialize TSDF volume by projecting depth map into the volume. Then we use volumetric shape-pose optimization (see Sec.~\ref{subsec:shape_opt}) to estimate initial shape parameters $\boldsymbol{\beta}_0$ and skeletal pose $\boldsymbol{\theta}_0$. After that, we initialize the double node graph using the on-body node graph and initial pose and shape as shown in Fig.~\ref{fig:eval_hieGraph}(a)(bottom). We extract a triangle mesh from the volume using Marching Cube algorithm~\cite{lorensen1987marching} and sample additional \emph{far-body nodes}. These nodes are used to parameterize non-rigid deformations far from inner body shape.

\begin{figure*}
	\begin{center}
		\includegraphics[width=1.0\linewidth]{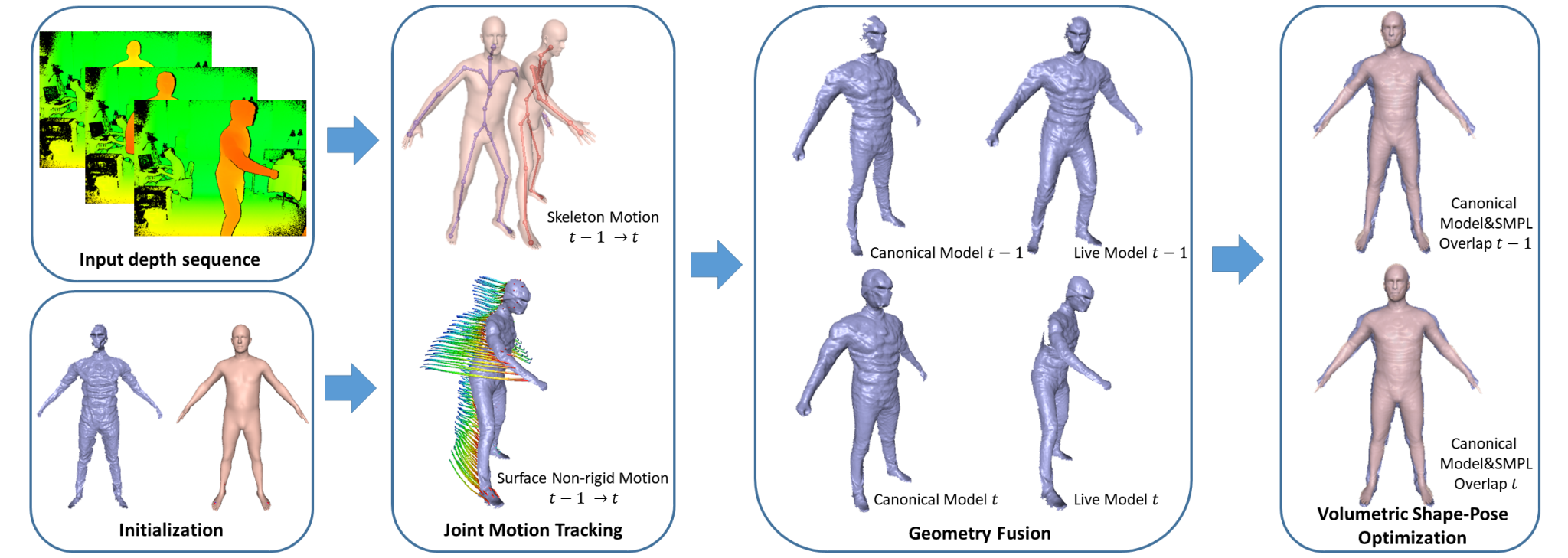}
	\end{center}
	\caption{Our system pipeline. We first initialize our system using the first depth frame (Sec.~\ref{subsec:init}). Then for each frame, we sequentially perform the next 3 steps: joint motion tracking ( Sec.~\ref{sec:tracking}), geometric fusion (Sec.~\ref{subsec:geometry_fusion}) and volumetric shape-pose optimization (Sec.~\ref{subsec:shape_opt}).}\vspace{-6pt}
	\label{fig:overview}
\end{figure*}

\subsection{Main Pipeline}
The main challenge to adopt SMPL in our pipeline is that initially the incomplete outer surface leads to difficult model fitting. Our solution is to continuously update the shape and pose in the canonical frame when more geometry is fused. Therefore, we propose a pipeline that executes \emph{joint motion tracking}, \emph{geometric fusion} and \emph{volumetric shape-pose optimization} sequentially (Fig.~\ref{fig:overview}). We briefly introduce the main components of the pipeline below:

\noindent\textbf{Joint Motion tracking} Given the current estimated parameters of body shape, we jointly optimize pose and the non-rigid deformations defined by the double node graph (Sec.~\ref{sec:tracking}). For the on-body nodes, we constrain the non-rigid deformations of them to follow skeletal motions. The far-body nodes are also optimized in the process but are not constrained by the skeleton.

\noindent\textbf{Geometric fusion} Similar to previous work~\cite{newcombe2015dynamic}, we non-rigidly integrate depth observation of multiple frames in a reference volume (Sec.~\ref{subsec:geometry_fusion}). We also explicitly detect collided voxels to avoid erroneously fused geometry~\cite{guo2017real}.

\noindent\textbf{Volumetric shape-pose optimization} After geometric fusion, the surface in the canonical frame gets more complete. We directly optimize the body shape and pose by using the fused signed distance field (Sec.~\ref{subsec:shape_opt}) This step is very efficient because it does not require finding correspondences.

\section{Joint Motion Tracking}
\label{sec:tracking}
There are two parameterizations in our motion tracking component, skeletal motions and non-rigid node deformations. Similar to the previous work~\cite{tao2017BodyFusion}, we adopt a binding term that constrains both motions to be consistent. Different from~\cite{tao2017BodyFusion}, we only enforce the binding term on on-body nodes to penalize non-articulated motions on on-body nodes. In contrast, far-body nodes have independent non-rigid deformations which are regularized to move like other nodes in the same graph structure. Besides geometric regularization, we also follow previous work~\cite{Bogo2016keepsmpl} to use a statistic pose prior to prevent unnatural poses. The energy of joint optimization is then
\begin{equation} \label{eqn:tracking_energy}
E_{\mathrm{mot}} = \lambda_{\mathrm{data}}E_{\mathrm{data}} + \lambda_{\mathrm{bind}}E_{\mathrm{bind}} + \lambda_{\mathrm{reg}}E_{\mathrm{reg}} + \lambda_{\mathrm{pri}}E_{\mathrm{pri}},
\end{equation}
where $E_{\mathrm{data}}$, $E_{\mathrm{bind}}$, $E_{\mathrm{reg}}$ and $E_{\mathrm{prior}}$ are energies of data, binding, regularization and pose prior term respectively.

\noindent \textbf{Data Term} The data term measures the fitting between the reconstructed double layer surface and depth map:
\begin{equation} \label{eqn:tracking_data_energy}
\begin{split}
E_{\mathrm{data}} = \sum_{(v_c, u)\in\mathcal{P}} {\tau_1(\mathbf{v}_c)*\psi(\tilde{\mathbf{n}}_{v_c}^{\mathrm{T}}(\tilde{\mathbf{v}}_c-\mathbf{u})) +} \\ {(\tau_2(\mathbf{v}_c)+\tau_3(\mathbf{v}_c))*\psi(\hat{\mathbf{n}}_{v_c}^{\mathrm{T}}(\hat{\mathbf{v}}_c-\mathbf{u}))},
\end{split}
\end{equation}
where $\mathcal{P}$ is the correspondence set; $\psi(\cdot)$ is the robust Geman-McClure penalty function; $(\mathbf{v}_c, \mathbf{u})$ is a correspondence pair; $\mathbf{u}$ is a sampled point on the depth map and its closest point $\mathbf{v}_c$ can be on either the body shape or fused surface. Correspondences on the body shape enable fast and robust tracking performance. $\tau_1({\mathbf{v}_c})$, $\tau_2({\mathbf{v}_c})$ and $\tau_3({\mathbf{v}_c})$ are correspondence indicator functions: $\tau_1({\mathbf{v}_c})$ equals to $1$ only if $\mathbf{v}_c$ is on the fused surface; $\tau_2({\mathbf{v}_c})$ equals to $1$ when $\mathbf{v}_c$ is on the body shape; $\tau_3({\mathbf{v}_c})$ equals to $1$ when $\mathbf{v}_c$ is on the fused surface and its 4 nearest nodes (knn-nodes) of $\mathbf{v}_c$ are all on-body nodes. $\tilde{\mathbf{v}}_c$ and $\tilde{\mathbf{n}}_{v_c}$ are the vertex position and normal warped by its knn-nodes using dual quaternion blending and defined as
\begin{equation} \label{eqn:dq_blend}
\mathbf{T}(\mathbf{v}_c) = SE3(\!\!\sum_{k\in\mathcal{N}(v_c)}\omega(k,v_c)\,\mathbf{dq}_k),
\end{equation}
where $\mathbf{dq}_j$ is the dual quaternion of $j$th node; $SE3(\cdot)$ maps a dual quaternion to $\mathbf{SE}$(3) space; $\mathcal{N}(v_c)$ represents a set of node neighbors of $\mathbf{v}_c$; $\omega(k, v_c)=\mathrm{exp}(-\lVert\mathbf{v}_c-\mathbf{x}_k\rVert^2_2/(2r_k^2))$ is the influence weight of the $k$th node $\mathbf{x}_k$ to $\mathbf{v}_c$; we set the influence radius $r_k=0.075\textrm{m}$ for all nodes.
$\hat{\mathbf{v}}_c$ and $\hat{\mathbf{n}_{v_c}}$ are the vertex position and its normal skinned by skeleton motions using linear blend skinning (LBS) and defined as
\begin{equation} \label{eqn:skeletal_lbs}
\begin{split}
\mathbf{G}(\mathbf{v}_c) &= \sum_{i\in\mathcal{B}}\: w_{i,v_c} \:\mathbf{G}_i, \\
\mathbf{G}_i &= \prod_{k\in\mathcal{K}_i} \mathrm{exp}(\theta_k\hat{\xi}_k),
\end{split}
\end{equation}
where $\mathcal{B}$ is index set of bones; $\mathbf{G}_i$ is the cascaded rigid transformation of $i$th bone; $w_{i,v_c}$ is the skinning weight associated with $k$th bone and point $\mathbf{v}_c$; $\mathcal{K}_i$ is parent indices of $i$th bone in the backward kinematic chain; $\mathrm{exp}(\theta_k\hat{\xi}_k)$ is the exponential map of the twist associated with $k$th bone. Note that the skinning weights of $\mathbf{v}_c$ is given by the weighted average of the skinning weights of its knn-nodes.

For each $\mathbf{u}$ on the depth map, we search for two types of correspondences on our double layer surface: $\mathbf{v}_t$ on the body shape and $\mathbf{v}_s$ on the fused surface. We choose the one that maximizes the following metric based on Euclidean distance and normal affinity
\begin{equation} \label{eqn:corresp_selection}
c = \underset{i\in\{t,s\}}{\mathrm{argmax}}\left(\left(1-\frac{\lVert\mathbf{v}_i-\mathbf{u}\rVert_2}{\delta_{\mathrm{max}}}\right)^2\!\! + \mu\,\tilde{\mathbf{n}}_{v_i}^{\mathrm{T}}\mathbf{n}_{u}\right),
\end{equation}
where we choose $\mu=0.2$; we set $\delta_{\mathrm{max}}=0.1\textrm{m}$ as the maximum radius used to search correspondences. We adopt two strategies for correspondence searching. To find correspondences between the depth map and the fused surface, we project the fused surface to 2D and then find correspondences within a local search window. For correspondences between the depth map and the body shape, we first find the nearest on-body node and then search for the nearest vertex around it. We eliminate the correspondences with distance bigger than $\delta_{max}$. These two methods are efficient for real-time performance and avoid building complex space partitioning data structure on GPU.
The binding term attaches on-body nodes to their nearest bones and helps to produce articulated deformations on the body. It is defined as

\begin{equation} \label{eqn:tracking_bind_energy}
E_{\mathrm{binding}} = \sum_{i\in\mathcal{L}_s} \lVert \mathbf{T}(\mathbf{x}_i)\mathbf{x}_i - \hat{\mathbf{x}}_i \rVert^2_2,
\end{equation}
where $\mathcal{L}_s$ is the index set of on-body nodes. $\hat{\mathbf{x}}_i$ is the node position skinned by LBS as defined in Eqn.~\ref{eqn:skeletal_lbs}.

\noindent \textbf{Regularization Term} The graph regularization is defined on all of the graph edges. This term is used to produce locally as-rigid-as-possible deformations. For on-body node graph, we decrease the effects of this regularization around joint regions by comparing the skinning weight vector of neighboring nodes as in ~\cite{tao2017BodyFusion}. This term is then defined as
\begin{equation}
E_{\mathrm{reg}} = \sum_i \sum_{j\in\mathcal{N}(i)} \rho(\lVert W_i-W_j\rVert^2_2)\: \lVert\mathbf{T}_i\mathbf{x}_j-\mathbf{T}_j\mathbf{x}_j\rVert^2_2
\end{equation}
where $\mathbf{T}_i$ and $\mathbf{T}_j$ are transformation associated with $i$th and $j$th nodes; $W_i$ and $W_j$ are skinning weight vectors of these two nodes respectively; $\rho(\cdot)$ is the Huber weight function in ~\cite{tao2017BodyFusion}.
Around joint regions, if two neighbor nodes are on different body parts, the difference of the skinning weight vectors is large, and thus $\rho(\cdot)$ will decrease the effect of the regularization. This will help to produce articulated deformations of on-body node graph. For far-body node graph, we construct its regularization term similar to ~\cite{newcombe2015dynamic}.

\noindent \textbf{Pose Prior Term} Similar to~\cite{Bogo2016keepsmpl}, we include a pose prior penalizing the unnatural poses. It is defined as
\begin{equation}
E_{\mathrm{prior}} = -\log \Big(\sum_j \omega_j N(\boldsymbol{\theta};\mu_j,\delta_j) \Big).
\end{equation}
This is formulated as a Gaussian Mixture Model (GMM), where $\omega_j$, $\mu_j$ and $\delta_j$ is the mixture weight, the mean and the variance of $j$th Gaussian model.

We solve the optimization problem (Eqn.~\ref{eqn:tracking_energy}) using Iterative Closest Point (ICP) method. First we build a correspondence set $\mathcal{P}$ using the latest motion parameters; then we solve the non-linear least squares using Gauss-Newton method. We use a twist representation for both the bone and node transformations. Within each iteration of Gauss-Newton procedure, the transformations are approximated using one-order Taylor expansion around the latest values. Then we solve the resulting linear system using a custom designed highly efficient preconditioned conjugate gradient (PCG) solver on GPU~\cite{guo2017real,dou2016fusion4d}.

\begin{figure}
	\centering
	\includegraphics[width=1.0\linewidth]{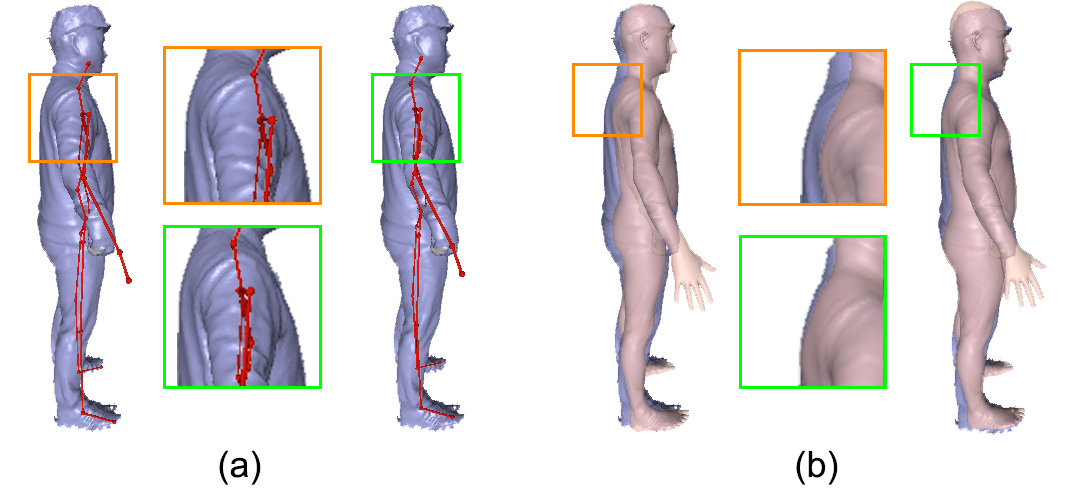}
	\caption{Illustration of volumetric shape-pose optimization. (a) skeleton embedding results before and after optimization. (b) shape-mesh overlap before and after optimization.}
	\label{fig:eval_shape_optm_illu} \vspace{4pt}
\end{figure}

\section{Volumetric Fusion \& Optimization} \label{sec:volumetric_fusion_opt}
\subsection{Geometric Fusion} \label{subsec:geometry_fusion}
Similar to the previous non-rigid fusion works~\cite{newcombe2015dynamic,innmann2016volume,guo2017real}, we integrate the depth information into a reference volume. First, the voxels in the reference volume are warped to live frame according to current non-rigid warp field. Then, we calculate the PSDF value of each valid voxel and use it to update their TSDF values. We follow the work~\cite{guo2017real} to cope with collided voxels in live frame to prevent erroneous fusion results caused by collisions. 

\subsection{Volumetric Shape-Pose Optimization} \label{subsec:shape_opt}
After the non-rigid fusion, we have an updated surface in the canonical volume with more complete geometry. Since the initial shape and pose parameters $(\boldsymbol{\beta}_0,\boldsymbol{\theta}_0)$ may not fit well with the new observation in the volume, as shown in Fig.\ref{fig:eval_shape_optm_illu}(a), we propose a novel algorithm that can efficiently optimize both of the shape parameters and initial embedding pose jointly in the canonical volume. The formulation of the energy is then
\begin{equation} \label{eqn:shape_energy}
E_{\mathrm{shape}} = E_{\mathrm{sdata}} + E_{\mathrm{sreg}} + E_{\mathrm{pri}},
\end{equation}
where $E_\mathrm{sdata}$ measures misalignment error in the reference volume; $E_{\mathrm{sreg}}$ is a temporal constraint that makes the new shape and poses parameters consistent with the previous ones. $E_{\mathrm{pri}}$ is the same as in Eqn.~\ref{eqn:tracking_energy} to prevent unnatural poses. The novel volumetric data term is defined as
\begin{equation} \label{eqn:shape_data_energy}
E_{\mathrm{sdata}}(\boldsymbol{\beta},\boldsymbol{\theta}) = \sum_{\bar{\mathbf{v}}\in\bar{\mathbf{T}}} \psi ( \mathbf{D}(W(T(\bar{\mathbf{v}};\boldsymbol{\beta},\boldsymbol{\theta});J(\boldsymbol{\beta}),\boldsymbol{\theta}) )),
\end{equation}
where $\mathbf{D}(\cdot)$ is a bilinear sampling function that takes a point in the canonical volume and returns interpolated TSDF. Note that $\mathbf{D}(\cdot)$ returns valid distance values only when the knn-nodes of the given point are all on-body nodes; otherwise $\mathbf{D}(\cdot)$ returns 0. This prevents the body shape from incorrectly fitting exterior objects, e.g., the backpack a performer is wearing. $\mathbf{v}=T(\bar{\mathbf{v}};\boldsymbol{\beta},\boldsymbol{\theta})$ modifies $\bar{\mathbf{v}}$ by shape blend shape and pose blend shape; $W(\mathbf{v};J(\boldsymbol{\beta},\boldsymbol{\theta}),\boldsymbol{\theta})$ deforms $\mathbf{v}$ using linear blend skinning. The temporal regularization is defined as
\begin{equation} \label{eqn:shape_reg}
E_{\mathrm{sreg}}(\boldsymbol{\beta},\boldsymbol{\theta},\boldsymbol{\beta}',\boldsymbol{\theta}') = \gamma_1 \lVert \boldsymbol{\beta}-\boldsymbol{\beta}' \rVert^2_2 + \gamma_2 \lVert \boldsymbol{\theta}-\boldsymbol{\theta}' \rVert^2_2.
\end{equation}
This term prevents the optimized shape and pose parameters $(\boldsymbol{\beta},\boldsymbol{\theta})$ from deviating the ones $(\boldsymbol{\beta}',\boldsymbol{\theta}')$ of the previous frame.

Note that $T(\bar{\mathbf{v}};\boldsymbol{\beta},\boldsymbol{\theta})$ includes both the pose and shape parameters, which makes $W(\mathbf{v};J(\boldsymbol{\beta},\boldsymbol{\theta}),\boldsymbol{\theta})$ a non-linear function. We find that generally the pose blend shape $B_p(\boldsymbol{\theta})$ in $T(\bar{\mathbf{v}};\boldsymbol{\beta},\boldsymbol{\theta})$ contributes much less to the modified body shape compared with the shape blend shape. Therefore we ignore the pose blend shape in $T(\bar{\mathbf{v}};\boldsymbol{\beta},\boldsymbol{\theta})$, and the resulting skinning formulation $W(T(\bar{\mathbf{v}};\boldsymbol{\beta});J(\boldsymbol{\beta},\boldsymbol{\theta}),\boldsymbol{\theta})$ becomes a linear function of $(\boldsymbol{\beta},\boldsymbol{\theta})$. This will generate a better energy landscape for the sampling based energy (Eqn.~\ref{eqn:shape_data_energy}) and make the convergence faster. Then we solve the resulting energy using the same GPU-based Gauss-Newton solver as in Sec.~\ref{sec:tracking}. At last, we update the body shape and pose that embedded into the canonical frame and recalculate the motion field and the skeleton motions. After more surface observation is fused into the TSDF volume, the body shape and canonical body pose get more accurate.  (Fig.~\ref{fig:eval_shape_optm_illu}(b)).

\begin{figure*}
	\begin{center}
		\includegraphics[width=1.0\linewidth]{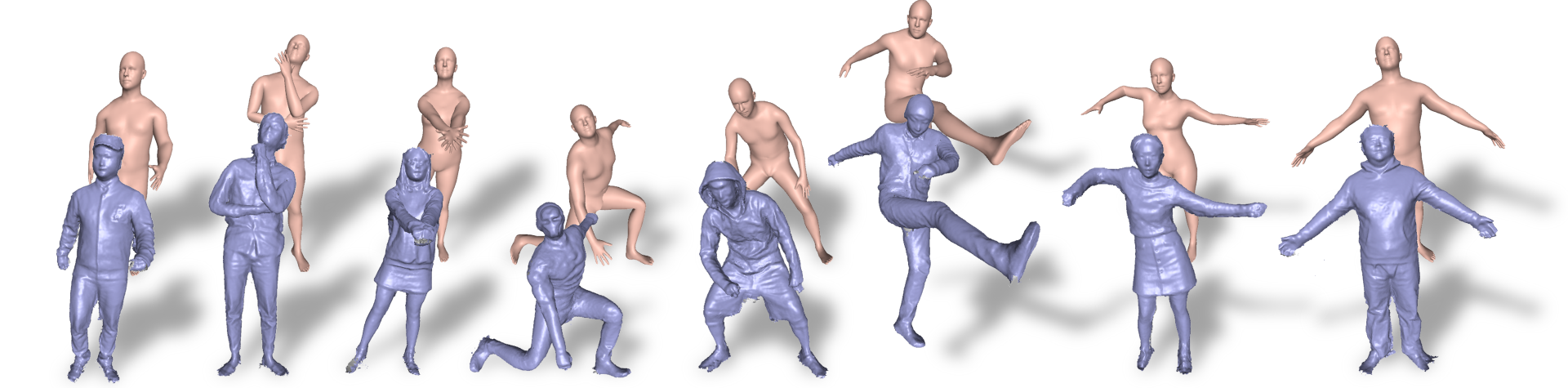}
	\end{center}\vspace{-10pt}
	\caption{ Example results reconstructed by our system. }\vspace{-10pt}
	\label{fig:results}
\end{figure*}

\section{Results}
\label{sec:results}
In this section, we first report the performance and the main parameters of the system. Then we compare with previous state-of-the-art methods qualitatively and quantitatively. We also evaluate each of our main contributions.
In Fig.~\ref{fig:results}, we demonstrate the results of our system. Note the various shapes, challenging motions and different types of cloth of the loop closed model that we can reconstructed. 

\subsection{Performance} \label{subsec:perf}
DoubleFusion runs in real-time (running at 32ms per frame). The entire pipeline is implemented on one NVIDIA TITAN X GPU. Executing 6 ICP iterations, the joint motion tracking takes 21 ms. The geometric fusion takes 6 ms and volumetric shape-pose optimization takes 3 ms. Prior to the joint motion tracking, we process the input depth frame using bilateral filtering, boundary outlier and floor plane removal. After volumetric shape-pose optimization, a triangulated mesh is extracted, non-rigidly transformed into camera coordinates and rendered on  the frame. These two parts run asynchronously with the main pipeline, and runtime overhead is negligible with less than 1 ms. For all of our experiments, we choose $\lambda_{\mathrm{data}}=1.0$, $\lambda_{\mathrm{bind}}= 1.0$, $\lambda_{\mathrm{reg}}=5.0$ and $\lambda_{\mathrm{pri}}=0.01$. For each vertex, we use its 4 nearest neighbors for warping; for each node, we use its 8 nearest neighbors to construct the node graph. The size of the voxel is set to 4 mm in each dimension.


\subsection{Evaluation} \label{subsec:evaluation}
\noindent \textbf{Double Node Graph.} We evaluate the proposed double node graph in Fig.~\ref{fig:eval_hieGraph}. The standard node graph construction scheme ~\cite{Sumner2007embededed} uniformly samples all the nodes on the fused outer surface. The lack of semantic information results in wrong connections (connection between two legs) and erroneous fusion results as shown in Fig.~\ref{fig:eval_hieGraph}(b). Using the on-body node graph alone is limited to capturing relatively tight clothing (e.g. the incomplete geometry of the backpack in Fig.~\ref{fig:eval_hieGraph}(c)) since it is out of the control area of on-body node graph. By using the proposed double node graph (Fig.~\ref{fig:eval_hieGraph}(d)), we can get clean and complete results.

\begin{figure}
	\centering
	\includegraphics[width=1.0\linewidth]{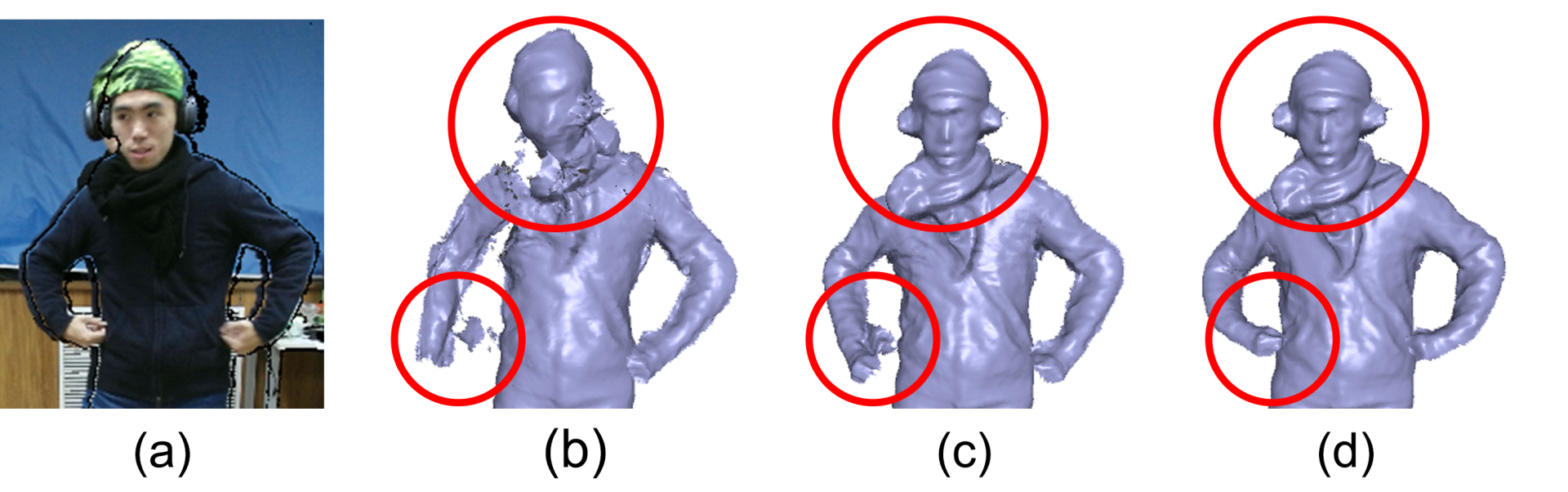}
	\caption{Evaluation of joint motion tracking. (a) reference color image. (b) results only using  correspondences on body for skeleton tracking, without non-rigid registration; (c) searching correspondences on both body and fused surface for skeleton tracking, without non-rigid registration; (d) using full energy terms.}\vspace{2pt}
	\label{fig:smpl_term_eval}
\end{figure}

\begin{figure}
	\centering
	\includegraphics[width=1.0\linewidth]{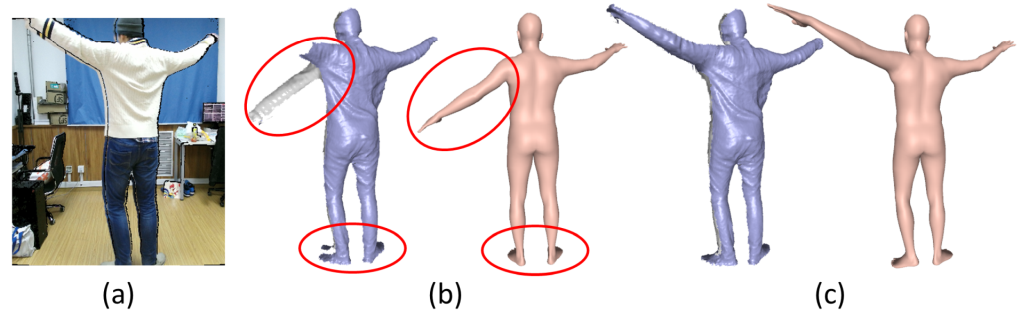}
	\caption{Evaluation of on-body correspondences. (a) reference color image (b) results only using fused surface for tracking. (c) results using both body and fused surface for tracking.}\vspace{4pt}
	\label{fig:eval_smplTerm}
\end{figure}

\noindent \textbf{Joint motion tracking.} In Fig.~\ref{fig:smpl_term_eval}, we evaluate different components of the joint motion tracking step qualitatively. We eliminate non-rigid registration in Fig.~\ref{fig:smpl_term_eval}(b) and (c). In Fig.~\ref{fig:smpl_term_eval}(b), we only use correspondences on the body shape by setting $\tau_1(\mathbf{v}_c)\equiv0, \tau_3(\mathbf{v}_c)\equiv0$ in Eqn.~\ref{eqn:tracking_data_energy}. It shows that without detailed surface and non-rigid registration, although an approximate pose can be tracked, the fused surface is noisy and erroneous; In Fig.~\ref{fig:smpl_term_eval}(c), we use correspondences on both body shape and fused surface by setting $\tau_1(\mathbf{v}_c)\equiv0$, the pose and fused surface get better but still contain artifacts. Only using all the energy terms we can get accurate pose and fusion results as shown in Fig.~\ref{fig:smpl_term_eval}(d). We also evaluate the on-body correspondences separately in Fig.~\ref{fig:eval_smplTerm}. Only using fused surface for tracking will easily get failed: will quickly fail when the left arm reappears with large motion in the scene due to the lack of surface geometry as shown in Fig.~\ref{fig:eval_smplTerm}(b). Using both surface and body shape for tracking will generate more plausible results as shown in Fig.~\ref{fig:eval_smplTerm}(c).

\begin{figure}
	\centering
	\includegraphics[width=0.95\linewidth]{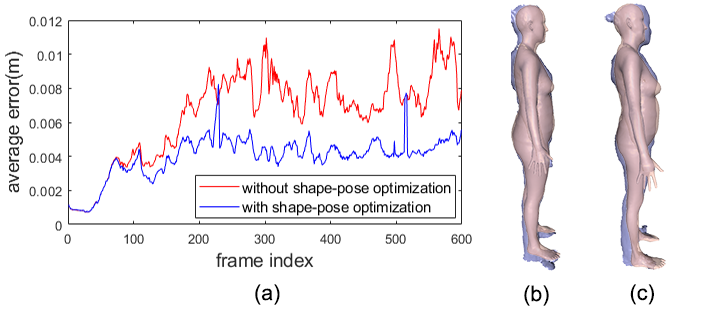}
	\caption{Evaluation of volumetric shape-pose optimization using non-rigid tracking accuracy. (a) average tracking error per frame, (b) reconstructed shape-mesh overlap with optimization, (c) reconstructed shape-mesh overlap without optimization.}\vspace{6pt}
	\label{fig:eval_shape_optm_quality}
\end{figure}

\begin{figure}
	\centering
	\includegraphics[width=0.95\linewidth]{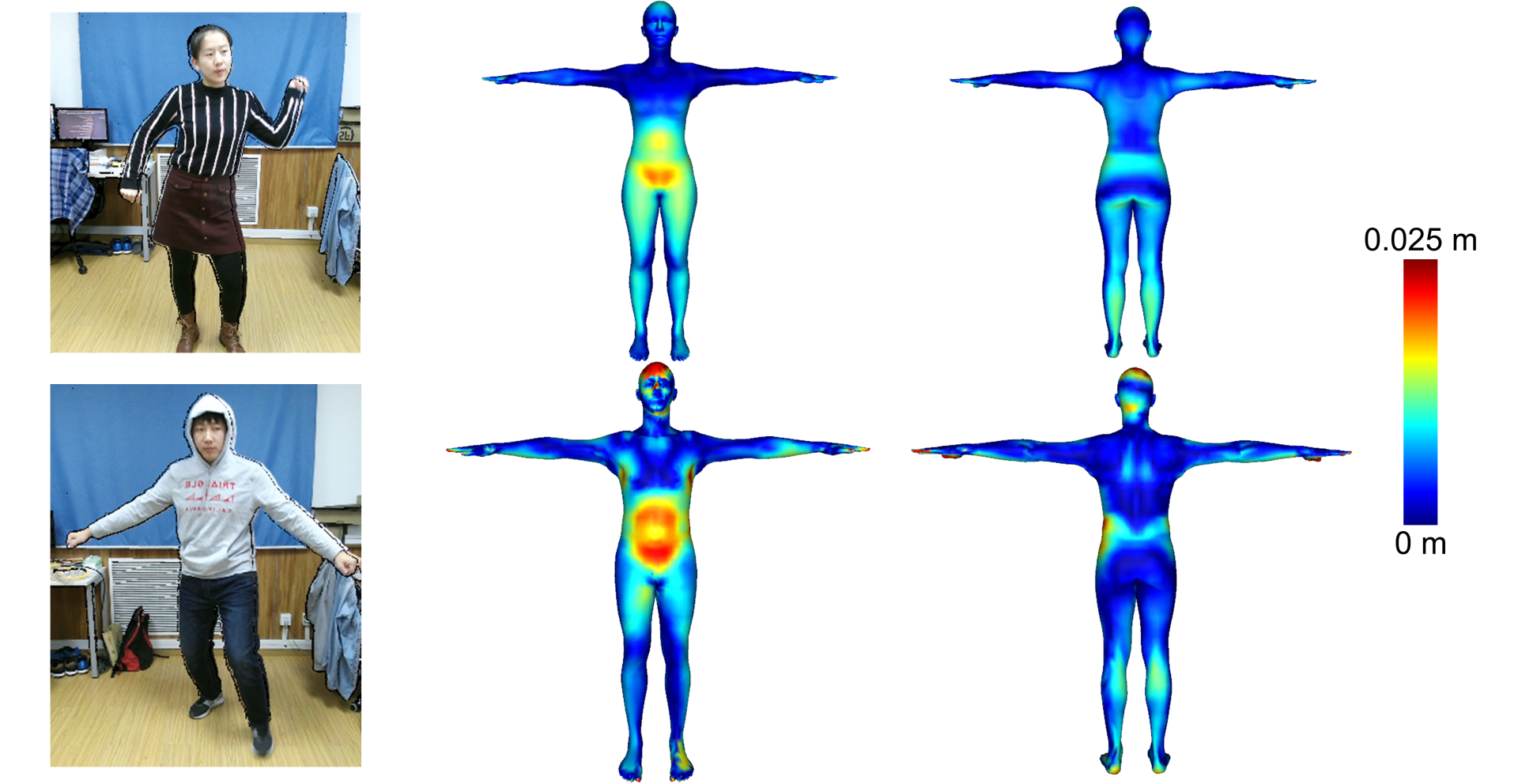}
	\caption{Per-vertex error of the reconstructed body shapes. }\vspace{5pt}
	\label{fig:eval_shape_optm_quantity_vis}
\end{figure}

\begin{figure}
	\centering
	\includegraphics[width=0.95\linewidth]{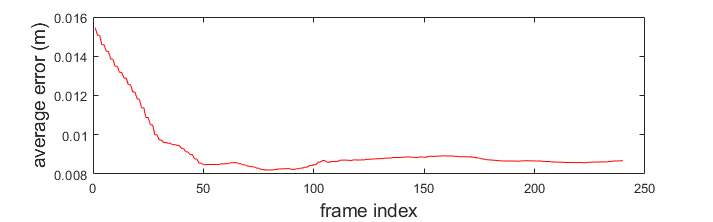}
	\caption{Evaluation of the body shape estimation accuracy of our online shape-pose optimization method. }\vspace{5pt}
	\label{fig:eval_shape_optm_quantity_curve}
\end{figure}

\noindent \textbf{Volumetric shape-pose optimization.} We evaluate volumetric shape-pose optimization both qualitatively and quantitatively. To evaluate non-rigid tracking accuracy, in Fig.~\ref{fig:eval_shape_optm_quality}, we use a public 4D sequence. We first render a single view depth sequence and then perform reconstruction using our system with/without optimization. The per-frame tracking error is calculated by averaging the point to plane error from the fused surface to the ground truth. We get better non-rigid tracking accuracy by using the optimization as shown in Fig.~\ref{fig:eval_shape_optm_quality}(a), and (b-c) demonstrates the reconstructed shape-mesh overlap with and without optimization. In Fig.~\ref{fig:eval_shape_optm_quantity_vis} and Fig.~\ref{fig:eval_shape_optm_quantity_curve}, we evaluate the accuracy of our reconstructed shape. We obtain ground truth undressed shape using laser scanner. Then we capture the same subject with clothing using DoubleFusion. As shown in Fig.~\ref{fig:eval_shape_optm_quantity_vis}, our reconstructed body shapes are plausible even though the subjects are dressed. Fig.~\ref{fig:eval_shape_optm_quantity_curve} shows the average shape reconstruction error along the sequence. 

\subsection{Comparison}
\begin{figure}
	\centering
	\includegraphics[width=0.95\linewidth]{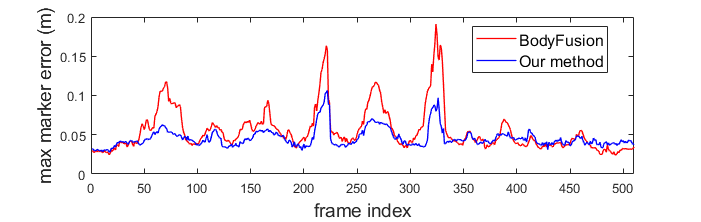}
	\caption{ Comparison of tracking accuracy on sequence "szq".}\vspace{5pt}
	\label{fig:quant_comparison_curve}
\end{figure}

\vspace{-.1cm}
\label{subsec:comparison}
We compare our tracking accuracy with BodyFusion~\cite{tao2017BodyFusion} using their public vicon dataset.
DoubleFusion obtains smaller per-frame max error (Fig.~\ref{fig:quant_comparison_curve}), and smaller average error (Tab.~\ref{tab:quant_comparison_value}), especially during fast motions.

\begin{table}
	\vspace{5pt}
	\setlength{\abovecaptionskip}{0pt}
	\begin{center}
		\begin{tabular}{|l|c|r|}
			\hline
			Method & BodyFusion\cite{tao2017BodyFusion} & Ours \\
			\hline
			Maximum Error (m) & 0.0554 & 0.0458 \\
			Average Error (m) & 0.0277 & 0.0221 \\
			\hline
		\end{tabular}
	\end{center}\vspace{-6pt}
	\caption{ Average numerical errors on the entire sequence.}\vspace{7pt}
	\label{tab:quant_comparison_value}
\end{table}

\begin{figure}
	\centering
	\includegraphics[width=0.95\linewidth]{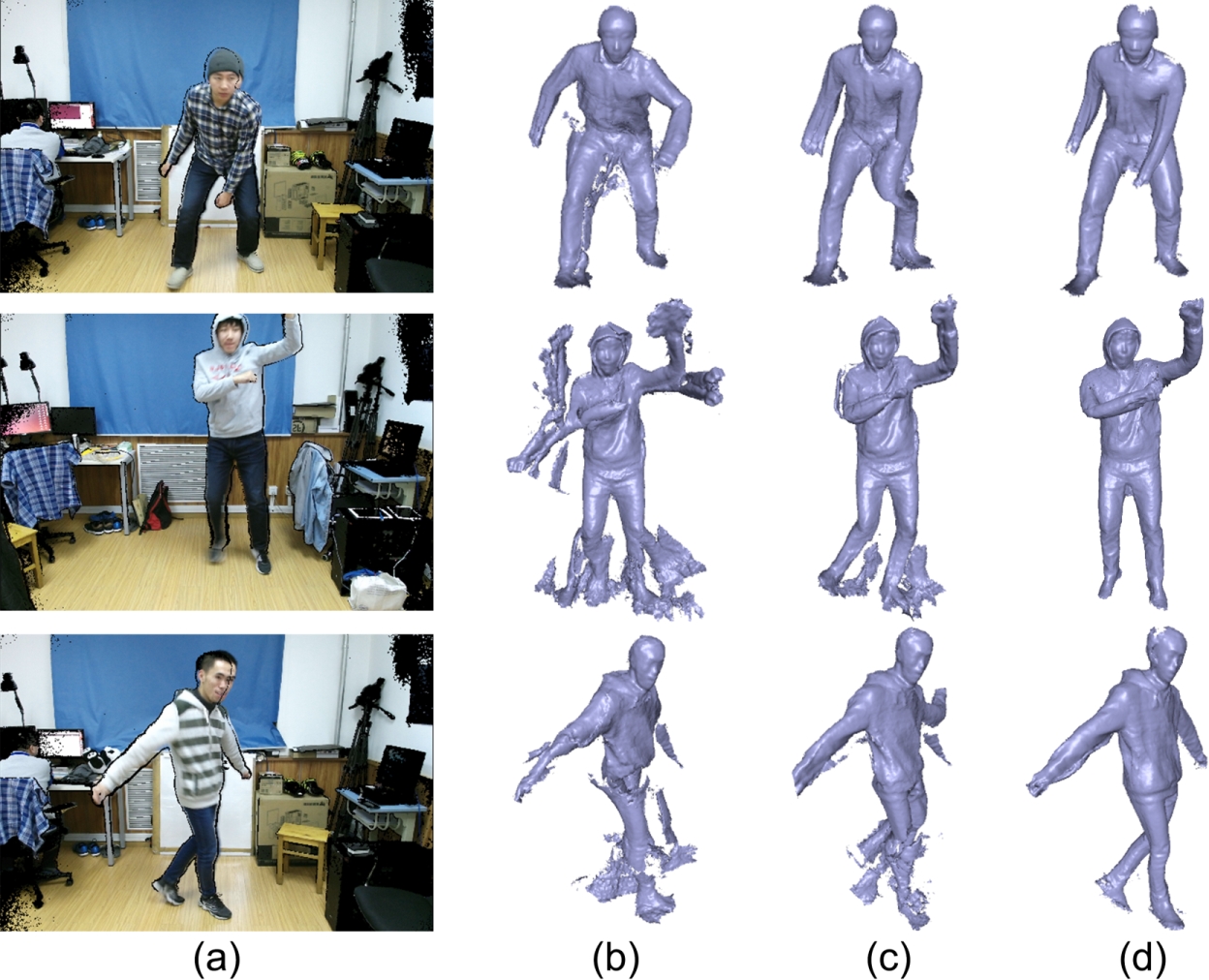}
	\caption{ Comparison. (a) reference color image. (b)(c)(d), results of DynamicFusion\cite{newcombe2015dynamic}, BodyFusion\cite{tao2017BodyFusion} and our method. }
	\label{fig:result_comparison}\vspace{4pt}
\end{figure}

We qualitatively compare our method with two real-time state-of-the-art methods~\cite{newcombe2015dynamic,tao2017BodyFusion}. \cite{newcombe2015dynamic} uses general non-rigid registration method without any prior, while~\cite{tao2017BodyFusion} takes advantage of a human skeletal constraint for better tracking ability. Fig.~\ref{fig:result_comparison} shows that our method achieves improved tracking and loop closure performance than other methods. Please see the supplementary video for more details.

\vspace{-12pt}
\section{Discussion}
\label{sec:discussion} \vspace{-0.2cm}
\noindent \textbf{Limitations} Our system tends to over-estimate body size when users wear thick clothing, and reconstruction of very wide cloth remains challenging. 
We cannot handle geometry separations of the outer surface, this could be addressed incorporating the key-volume update method in~\cite{dou2016fusion4d}. 
Our current system can not handle human-object interactions, which we plan to address in future work.

\noindent \textbf{Conclusion} In this paper, we have demonstrated the first method for real-time reconstruction of both clothing and inner body shape from a single depth sensor. Based on the proposed double surface representation, our system achieved better non-rigid tracking and surface loop closure performance than state-of-the-art methods. Moreover, the real-time reconstructed inner body shapes are visually plausible. We believe the robustness and accuracy of our approach will enable many applications, especially in AR/VR, gaming, entertainment and even virtual try-on as we also reconstruct the underlying body shape. For the first time, with DoubleFusion, users can easily digitize themselves. 

\noindent \textbf{Acknowledgements} This work is supported by the National key foundation for exploring scientific instrument of China No.2013YQ140517; NKBRP of China No.2014CB744201; the National NSF of China grant No.61522111, No.61531014, No.61233005; Shenzhen Peacock Plan KQTD20140630115140843; Changjiang Scholars and Innovative Research Team in University, No.IRT\_16R02; Google Faculty Research Award; the Okawa Foundation Research Grant; the U.S. Army Research Laboratory under contract W911NF-14- D-0005.



{\small
\bibliographystyle{ieee}
\bibliography{main_paper}
}

\end{document}